\documentclass{article}

\PassOptionsToPackage{numbers, compress}{natbib}

\usepackage[final]{neurips_2021}

\usepackage[utf8]{inputenc} 
\usepackage[T1]{fontenc}    
\usepackage{hyperref}       
\usepackage{url}            
\usepackage{booktabs}       
\usepackage{amsfonts}       
\usepackage{nicefrac}       
\usepackage{microtype}      
\usepackage[dvipsnames]{xcolor}         

\usepackage[small]{caption}
\usepackage{graphicx}
\usepackage{amsmath}
\usepackage{amsthm}
\usepackage{algorithm}
\usepackage{algorithmic}
\usepackage{microtype}
\usepackage{amssymb}
\usepackage{color}
\usepackage{subcaption}
\usepackage{xspace}
\usepackage{tabularx}
\usepackage{inconsolata}
\usepackage{multicol}
\usepackage{wrapfig}
\usepackage{makecell}
\usepackage{chngpage}
\usepackage{etoolbox}

\newtheorem{definition}{Definition}

\newcommand{\commentout}[1]{}
\usepackage[normalem]{ulem}
\usepackage{soul}

\newcommand{\fail}{\mbox{\sffamily{x}}}

\usepackage{marginnote}

\newtoggle{arxiv}
\toggletrue{arxiv}
\newcommand{\ifArxiv}[1]{\iftoggle{arxiv}{#1}{the full version of this paper \cite{pouget2021ranking}}}

\graphicspath{{./imgs/}}

\title{Ranking Policy Decisions}

\author{
  Hadrien Pouget\thanks{The work in this paper was done while at the University of Oxford.}\\
  University of Cambridge\\
  UK\\
  \texttt{pougeth@gmail.com}\\
  \And
  Hana Chockler\\
  causaLens\\
  and\\
  King's College London\\
  UK\\
  \texttt{hana@causalens.com}\\
  \texttt{hana.chockler@kcl.ac.uk}\\
  \AND
  Youcheng Sun\\
  Queen's University Belfast\\
  UK\\
  \texttt{youcheng.sun@qub.ac.uk}\\
  \And
  Daniel Kroening\thanks{The work in this paper was done prior to joining Amazon.}\\
  Amazon\\
  UK\\
  \texttt{daniel.kroening@magd.ox.ac.uk}
}

\begin{document}
\maketitle
\begin{abstract}
	Policies trained via Reinforcement Learning (RL)
	are often needlessly complex,
	making them difficult to analyse and interpret.
	In a run with $n$ time steps, a policy will make $n$ decisions on 
	actions to take; we conjecture that only
	a small subset of these decisions delivers 
	value over selecting a simple default action.
	Given a trained policy, we propose a novel black-box method 
	based on statistical fault localisation
	that ranks the states of the environment
	according to the importance of decisions made in those states.
	We argue that among other things, the ranked list of states can help explain and
	understand the policy. As the ranking method is statistical, a direct
	evaluation of its quality is hard. As a proxy for quality,
	we use the ranking to create new, simpler policies from the original ones
	by pruning decisions identified as unimportant
	(that is, replacing them by default actions)
	and measuring the impact on performance.
	Our experiments on a diverse set of standard benchmarks
	demonstrate that pruned policies can perform on a level comparable
	to the original policies.
	Conversely, we show that naive approaches for ranking policy decisions,
	e.g., ranking based on the frequency of visiting a state,
	do not result in high-performing pruned policies. 
	
\end{abstract}

\section{Introduction}

Reinforcement learning is a powerful 
method for training policies that
complete tasks in complex environments.
The policies produced are optimised to maximise 
the expected reward provided by the environment. While
performance is clearly an important goal, 
the reward typically does not capture the 
entire range of our preferences. By focusing solely on 
performance, we risk overlooking the demand for
models that are easier to analyse, 
predict and interpret~\cite{lewis2021deep}.
Our hypothesis is that many trained policies
are \emph{needlessly complex}, i.e., that there
exist alternative policies that
perform just as well or nearly as well
but that are significantly simpler.
This tension between performance and simplicity 
is central to the field of explainable AI (XAI), and 
machine learning as a whole~\cite{gunning2019darpa};
our method aims to help by highlighting the most important 
parts of a policy.

The starting point for our definition of ``simplicity'' is
the assumption that there exists a way to make a ``simple choice'',
that is, there is a 
simple default action
for the environment. We argue that this is the case
for many environments in which RL is applied: for example,
``repeat previous action'' is a straightforward default action
for navigation tasks. 

The key contribution of this paper is a novel method for
\emph{ranking policy decisions} according to their importance
relative to some goal. We argue that the ranked list of decisions
is already helpful in explaining how the policy operates.
We evaluate our ranking method by
using the ranking to simplify policies without compromising performance,
hence addressing one of the main hurdles for wide adoption
of deep RL: the high complexity of trained policies.
%

We produce a ranking
by scoring the states a policy visits. The rank reflects
the impact that replacing the policy's chosen action by
the default action has on a user-selected binary outcome, such as
``obtain more than X reward''.
It is intractable to compute this ranking precisely, owing to the
high complexity and the stochasticity of the environment and the policy, 
complex causal interactions between actions and their outcomes,
and the sheer size of the problem. 
%
Our work uses \emph{spectrum-based fault localisation}
(SBFL) techniques~\cite{naish2011model,wong2016survey}, borrowed from the software
testing domain, to compute an approximation of the ranking of policy decisions.
%
SBFL is an established technique
in program testing for ranking the parts of a program
source code text that are most likely to contain the
root cause of a bug. This ranking is computed by recording
the executions of a user-provided test suite. SBFL distinguishes
passing and failing executions; failing executions are those
that exhibit the bug. Intuitively, a program location is
more likely to be the root cause of the bug if it is visited
in failing executions but less (or not at all) in passing ones.
SBFL is a lightweight technique and its rankings are 
highly correlated with the 
location of the root cause of the bug~\cite{wong2016survey}.
We argue that SBFL is also a good fit for analysing complex RL policies.


Our method applies to RL policies in a black-box manner, and requires
no assumptions about the policy's training or representation. We evaluate
the quality of the ranking of the decisions by the proxy of creating new,
simpler policies (we call them ``pruned policies'') without retraining,
and then calculate the
reward achieved by these policies. Experiments with agents
for MiniGrid (a~more complex version of gridworlds)~\cite{gym_minigrid},
CartPole~\cite{DBLP:journals/corr/BrockmanCPSSTZ16} and
a number of Atari games~\cite{DBLP:journals/corr/BrockmanCPSSTZ16}
demonstrate that pruned policies maintain high performance (similar or
only slightly worse than that of the original policy) when
taking the default action in the majority of the states (often $90\%$
of the states).
As pruned policies are much easier to understand than the original policies,
we consider this an important step towards explainable RL. 
Pruning a given policy does not require re-training, and hence,
our procedure is relatively lightweight.
Furthermore,
the ranking of states by itself provides important insight into the importance
of particular decisions for the performance of the policy overall.


The code for reproducing our experiments is available on GitHub\footnote{\url{https://github.com/hadrien-pouget/Ranking-Policy-Decisions}. Experiments done at commit
c972414}, 
and further examples are provided on the project
website\footnote{\url{https://www.cprover.org/deepcover/neurips2021/}}.


\begin{figure}
    \begin{subfigure}[t]{0.4\linewidth}
        \centering
        \includegraphics[width=0.49\linewidth]{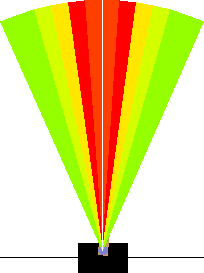}\hfill%
        \includegraphics[width=0.49\linewidth]{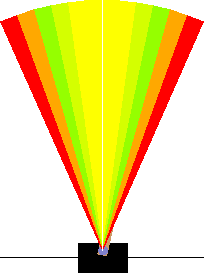}
        \caption{}
        \label{fig:vis_1}
    \end{subfigure}\hfill%
    \begin{subfigure}[t]{0.52\linewidth}
        \centering
        \hfill
        \includegraphics[width=0.47\linewidth]{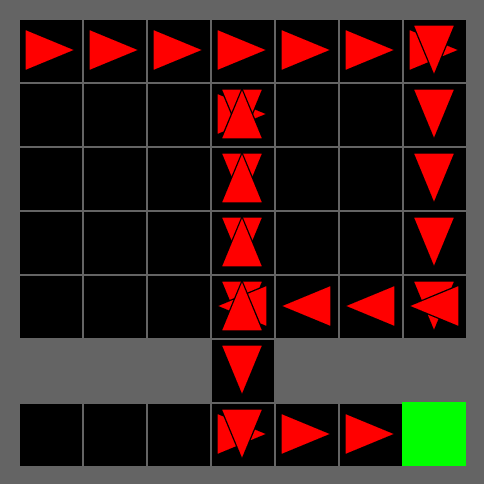}\hfill%
        \includegraphics[width=0.47\linewidth]{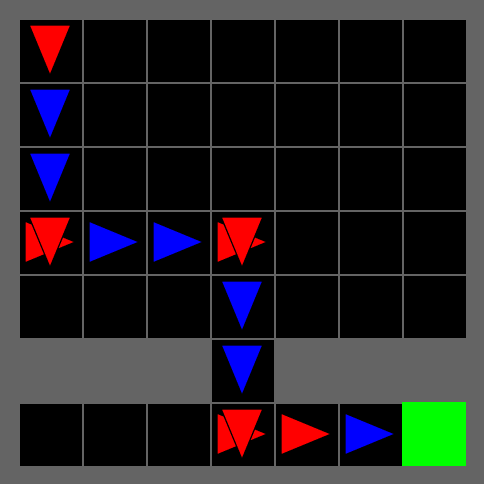}\hfill
        \caption{}
        \label{fig:vis_2}
    \end{subfigure}
    \caption{\textbf{(a)} CartPole, in a state where the cart and pole are moving rapidly.
        The heatmap represents 
        the frequency of appearance of the possible pole 
        angles (left) and the importance scores following SBFL (right). While
        it is more frequent for the pole to be 
        centred, SBFL successfully identifies that the policy's decisions are 
        more important when the pole is close to falling.
        \textbf{(b)} MiniGrid. Traces of executions with the original policy (left) and a \emph{pruned policy} (right). States in which 
        we take the default action are in blue. Both policies succeed, but pruning unimportant actions simplifies the policy.}
\end{figure}

\section{Background}
\label{sec:background}

\subsection{Reinforcement learning (RL)}\label{sec:rl}

We use a standard reinforcement learning (RL) setup and assume that the reader is familiar with the basic concepts.
An \emph{environment} in RL is defined as a Markov decision process (MDP) and is 
denoted by $\langle S, A, P, R, \gamma, T\rangle$, where $S$ is the set of states, $A$ is the set of actions, $P$ is the transition function, $R$ is the reward function, $\gamma$ is the discount factor, and $T$ is the set of terminal states. An agent seeks to learn
a policy $\pi:S\rightarrow A$ that maximizes the total discounted reward. Starting from the initial state $s_0$ and given the policy $\pi$, the state-value function is the expected future discounted reward as follows:
\begin{equation}\label{eq:rlvalue}
V_{\pi}(s_0) = \mathbb{E}\left(\sum_{t=0}^{\infty}\gamma^tR(s_t,\pi(s_t),s_{t+1})\right)
\end{equation}
A policy $\pi: S \rightarrow A$ maps states to the actions taken in these states and may be  stochastic. We treat the policy as a black box, and hence make no further assumptions about $\pi$. 

\subsection{Spectrum-based fault localization (SBFL)}
\label{sec:SFL}

The reader is likely less familiar with spectrum-based fault localization (SBFL), as to the best
of our knowledge, it has not yet been used in RL. We therefore give a detailed description.
SBFL techniques~\cite{wong2016survey}
have been widely used as an efficient approach to aid in locating the
causes of failures in sequential programs. 
SBFL techniques rank program elements (say program statements)
based on their \emph{suspiciousness scores}, which are computed using correlation-based
measures. Intuitively, a program element is more suspicious if it appears in failed
executions more frequently than in correct executions, and the exact formulas differ between
the measures. Diagnosis of the faulty program can then be conducted by
manually examining the ranked list of elements
in descending order of their suspiciousness until the cause of the fault is found. It has
been shown that SBFL techniques perform well in complex programs \cite{abreu2009practical}.

SBFL techniques first execute the program under test using a \emph{test suite}.
A test suite comprises of a set of inputs and an expected output for each input. A test \emph{passes} when the output produced by the program under test matches the expected output given by the test suite, and has \emph{failed} otherwise.
In addition to the outcome of the test, SBFL techniques record the values of a set of Boolean flags that indicate whether a particular program element was executed by that test. 

The task of a fault localization tool is to compute a ranking of the program elements
based on the values of the Boolean flags recorded while executing the test suite.
Following the notation from~\cite{naish2011model}, the suspiciousness score of each
program statement $s$ is calculated from a set of parameters
$\langle a^s_\mathit{ep}, a^s_\mathit{ef}, a^s_\mathit{np}, a^s_\mathit{nf} \rangle$ that
give the number of times the statement $s$ is executed ($e$) or not executed~($n$) on passing ($p$)
and on failing ($f$) tests. For instance, $a^s_\mathit{ep}$ is the number of tests that passed and executed $s$.

Many measures have been proposed to calculate the suspicious scores of program elements.
In Equations~(\ref{eq:measures}a)$\sim$(\ref{eq:measures}d) we list a selection of popular and high-performing 
measures~\cite{abreu2007accuracy,zoltar,jones2005empirical,wong2007effective};
these are also the measures that we use in our ranking procedure.

\begin{subequations}\label{eq:measures}%
\begin{tabular}{@{\hspace{-0.2cm}}l@{}l}%
\begin{minipage}{6.5cm}%
\begin{equation}\text{Ochiai:}\,\,\, \frac{a_\mathit{ef}^s}{\sqrt{(a_\mathit{ef}^s+a_\mathit{nf}^s)(a_\mathit{ef}^s+a_\mathit{ep}^s)}}\label{eq:ochiai}\end{equation}
\end{minipage}
& \begin{minipage}{6.5cm}%
\begin{equation}
\hspace{1.2cm}\text{Tarantula:}\,\,\,  \displaystyle\frac{\frac{a_\mathit{ef}^s}{a_\mathit{ef}^s+a_\mathit{nf}^s}}{\frac{a_\mathit{ef}^s}{a_\mathit{ef}^s+a_\mathit{nf}^s}+\frac{a_\mathit{ep}^s}{a_\mathit{ep}^s+a_\mathit{np}^s}}\label{eq:tarantula}\end{equation}
\end{minipage}\\
\begin{minipage}{6.5cm}%
\begin{equation}\text{Zoltar:}\,\,\, \frac{a_\mathit{ef}^s}{a_\mathit{ef}^s+a_\mathit{nf}^s+a_\mathit{ep}^s+\frac{10000a_\mathit{nf}^sa_\mathit{ep}^s}{a_\mathit{ef}^{s}}}\end{equation}
\end{minipage}
& \begin{minipage}{6.5cm}%
\begin{equation}
\text{Wong-II:} \label{eq:wong-ii}\,\,\,   a_\mathit{ef}^s-a_\mathit{ep}^s
\end{equation}
\end{minipage}
\end{tabular}
\end{subequations}

\commentout{
\begin{subequations}\label{eq:measures}%
    \begin{equation}\text{Ochiai:}\,\,\, \frac{a_\mathit{ef}^s}{\sqrt{(a_\mathit{ef}^s+a_\mathit{nf}^s)(a_\mathit{ef}^s+a_\mathit{ep}^s)}}\label{eq:ochiai}\end{equation}

    \begin{equation}\text{Tarantula:}\,\,\,  \displaystyle\frac{\frac{a_\mathit{ef}^s}{a_\mathit{ef}^s+a_\mathit{nf}^s}}{\frac{a_\mathit{ef}^s}{a_\mathit{ef}^s+a_\mathit{nf}^s}+\frac{a_\mathit{ep}^s}{a_\mathit{ep}^s+a_\mathit{np}^s}}\label{eq:tarantula}\end{equation}
    
    \begin{equation}\text{Zoltar:}\,\,\, \frac{a_\mathit{ef}^s}{a_\mathit{ef}^s+a_\mathit{nf}^s+a_\mathit{ep}^s+\frac{10000a_\mathit{nf}^sa_\mathit{ep}^s}{a_\mathit{ef}^{s}}}\end{equation}

    \begin{equation}
    \text{Wong-II:} \label{eq:wong-ii}\,\,\,   a_\mathit{ef}^s-a_\mathit{ep}^s
    \end{equation}
\end{subequations}
}

SBFL-based tools present the list of program elements in descending order of their suspiciousness scores to the user. There is no single best measure for fault localization; different measures perform better on different types of programs, and it is best practice to use multiple measures \cite{lucia2014extended}. In our experiments, we combine the 
four measures listed above for this very reason.

While more sophisticated versions of SBFL exist \cite{DBLP:conf/ijcai/AbreuZG09,DBLP:conf/ecai/CardosoAFK16}, in this work we prefer to stick to the simpler approach, which was sufficient for producing notable results.


\section{Method: ranking policy decisions using SBFL}\label{sec:approach}

Inspired by the use of SBFL for localising the cause of a program's outcome,
we propose a new SBFL-based method to identify the states in which 
decisions made by an RL policy are most important for achieving its
objective.
Our method is modular and is composed of two phases:
(1)~generating mutant policies and  (2)~ranking states based on the importance.

\subsection{Definitions}\label{sec:def}

\paragraph{Executions of RL policies}

We apply the SBFL technique to a set of \emph{executions}
(sometimes called \emph{trajectories} in the literature) of a given RL policy $\pi$ with 
mutations. An \emph{execution} $\tau$ of $\pi$ describes a traversal of the agent through the environment MDP
using the RL policy~$\pi$ and is defined as a sequence of states $s_0, s_1, \ldots$ and actions $a_0, a_1, \ldots$,
where $s_0$ is an initial state and each subsequent state $s_{i+1}$ obtained from the previous state $s_i$ by performing
action~$a_i$, as chosen using $\pi(s_i)$. The last state must be a terminal state. As $\pi$ or the environment can be
stochastic, each execution of $\pi$ may result in a different sequence of actions and states, and hence in
a different $\tau$. The set of all possible executions is denoted by $\mathcal{T} $.
A \emph{decision} of a policy $\pi$ in a state $s$ is a pair $\langle s, \pi(s) \rangle$. Note that $\pi(s)$ is the learned probability distribution
from which an action in this state is obtained; in a deterministic policy, $\pi(s)$ is a single action for each $s$.

\paragraph{Passing and failing executions}

An execution is either successful or failed. We define the success of an execution as a (binary)
value of a given assertion on this execution. For example, the assertion can be that
the agent reaches its destination eventually, or that the reward of this execution is
not lower than $0.75$ of the maximal reward for $\pi$. The assertion induces a Boolean
function $C: \mathcal{T} \rightarrow \{0, 1\}$. We say that an execution~$\tau$ is a \emph{pass} if $C(\tau) = 1$, and is a
\emph{fail} otherwise. We use a binary condition for simplicity, as SBFL is 
designed to work with passing and failing executions. This condition 
can be relaxed \cite{DBLP:conf/ecai/CardosoAFK16}, and we plan to investigate 
generalising this in future work.

\paragraph{Mutant executions}

We use SBFL to understand the impact 
of replacing actions by a \emph{default action}.
The choice of the default action $d$ is context dependent and can
be configured by the user. For example, an obvious default
action for navigation is ``proceed in the same direction''.
The default action can in principle be as basic as a single action,
or as complex as a fully-fledged policy. In our experiments,
we evaluate two default actions. The first is ``repeat the previous action'', defined as:
\begin{equation}
d(s_0, \ldots, s_i, a_0, \ldots, a_{i-1}) = a_{i-1},
\end{equation}
and the second is ``take a random action''. For $A$ being the action space,
\begin{equation}
d(s_0, \ldots, s_i, a_0, \ldots, a_{i-1}) = a_{s_i} \sim \mathit{Unif}(A).
\end{equation}
Once $a_{s_i}$ has been sampled, it is not re-sampled if the state
$s_i$ is revisited; the same action is used.
Using ``take a random action'' as the default action can be useful in cases where there is no other obvious default action. 
However, we generally expect this to be a worse option than a default action tailored to the environment. As the choice of $d$ depends on the context and the user's goals, it is ultimately a user choice.

Using the default action, we create \emph{mutant executions},
in which the agent takes the default action $d$ whenever 
it is in one of the \emph{mutant states}. More formally, given a 
set of mutant states $S_M$, we act according to the policy $\pi_{S_M}$
defined as:
\begin{equation}\label{eq:lazypolicy}
\pi_{S_M}(s_0, \ldots, s_i, a_0, \ldots, a_{i-1}) = 
\begin{cases}
\pi(s_i) & s_i \notin S_M, \\ 
d(s_0, \ldots, s_i, a_0, \ldots, a_{i-1}) & s_i \in S_M.
\end{cases}
\end{equation}

Decisions made in states in which the default action is a very good option are deemed less important. 
In these states, not following the policy would be less consequential.


\subsection{Generating the test suite and mutant executions on-the-fly}
\label{sec:testsuite}

The na\"ive approach to generating a comprehensive suite of mutant executions for applying SBFL would be to
consider all possible sets of mutant states $S_M$---that is, we would need to consider all possible subsets of the state space~$S$. 
However, the state space of most RL environments is too large to enumerate, and enumerating all
possible subsets of $S$ is intractable even for simplistic environments.

We use two algorithmic techniques to address this problem: (a)~we generate
mutant executions \emph{on-the-fly}, and (b)~we use an abstraction function
$\alpha : S \rightarrow \hat S$ to map the full set of states $S$ to a smaller, less complex
set of abstract states $\hat S$. Examples of these are in the supplementary material,
and may for example include down-scaling or grey-scaling images that are input 
to the agent, or quantising a continuous state space. 
The set of possible mutations is then the set of subsets
of $\hat S$, instead of the set of subsets of~$S$. We then score the abstract states, rather than the full state space. The test suite of 
mutant executions produced 
this way for $\pi$ is denoted $\mathcal{T}(\pi) \subseteq \mathcal{T}$.

\paragraph{On-the-fly mutation}

We maintain a set $S_M \subseteq \hat S$ per execution. We begin each execution 
$\tau$ by initialising
the set $S_M$ with the empty set of states.
At each step of the execution, upon visiting a state~$s$, we check the current $S_M$.
If $\alpha(s) \not\in S_M$, we add $\alpha(s)$ to $S_M$ according to the predefined \emph{mutation rate}~$\mu$
(and take the default action); otherwise, we use $\pi$ to determine the action in
this state. In case we re-visit an (abstract) state, we maintain the previous decision of whether to
mutate state $s$ or not. This way, the states that are never visited in any of the
executions are never mutated; hence, we never consider ``useless mutations'' that mutate a state
that is never visited. We finish by marking $\tau$ as
pass or fail according to $C(\tau)$.
Note that a mutant execution may visit states not typically encountered
by~$\pi$, meaning that we are even able to rank states that are out of distribution. This is especially important when trying to 
understand how the policy behaves in parts of the environment it 
is unfamiliar with.

Overall, our algorithm has five (tunable) parameters: the size of the test suite~$|\mathcal{T}(\pi)|$,
the passing condition~$C$,
the default action~$d$, the mutation rate~$\mu$ and the abstraction function~$\alpha$.

SBFL benefits from a balanced test suite of passing and failing 
executions~\cite{wong2016survey}, a ratio largely determined
by the choices of $\mu$ and $C$. The choice of $C$ depends on the context.
In our experiments in Section~\ref{sec:prunedpol}, 
our goal was to find the states with highest
impact on the expected reward. We set the condition to be ``receive more than $X$ 
reward" for some $X \in \mathbb{R}$, and chose $X$ to yield a balanced 
suite (i.e. if there were too many failed runs, we lowered $X$). 
In most cases actions important for achieving at least $X$ reward
are important for maximising reward in general, so
we found that this worked well. In the cases where some actions 
are not important for achieving $X$, but later important for 
getting even higher rewards 
(e.g., states that only appear after achieving $X$ reward), 
we would not have considered them important. 
The choice of $\mu$ is also significant. If it is too large, executions
fail too often, and the behaviour in mutant executions is uninteresting.
If $\mu$ is too small, we do not mutate states
enough to learn anything, and in larger 
environments fail to mutate many of the states we encounter. 
In our experiments, we selected $\mu$ manually, but this could easily 
be automated by a parameter tuning algorithm.

\subsection{Computing the ranking of the policy decisions} 
\label{sec:rankings}

We now explain how to rank states according to the importance of policy decisions made 
in these states, with respect to satisfying the condition.
The ranking method is based on SBFL as described in Section~\ref{sec:SFL}.
We first create the test suite of mutant 
executions $\mathcal{T}(\pi)$ as described above.
We denote the set of all abstract states encountered when generating the test suite $S_{\mathcal{T}} \subseteq \hat S$; these are the states to which we 
assign scores. Any unvisited state is given the lowest possible score by default.

Similarly to SBFL for bug localisation, for each state $s\in S_{\mathcal{T}}$ we
calculate a vector $\langle a^s_\mathit{ep}, a^s_\mathit{ef}, a^s_\mathit{np}, 
a^s_\mathit{nf} \rangle$. We use this vector to track 
the number of times that $s$ was unmutated ($e$) or mutated ($n$)
on passing ($p$) and on failing ($f$) executions, and
we update these scores based on those executions in which the state was visited.
In other words, the vector keeps track of
success and failure of mutant executions based on whether an execution took the 
default action in $s$ or not.
For example, $a^s_{ep}$ is the number of passing executions that took the action $\pi(s)$ 
in the state $s$, and $a^s_{nf}$ is the number of failing executions that took the default action
in the state~$s$. 
\begin{definition}[Ranking]\label{def-rank}
Given an SBFL measure $m$ and a vector
$\langle a^s_\mathit{ep}, a^s_\mathit{ef}, a^s_\mathit{np}, a^s_\mathit{nf} \rangle$ for each (abstract)
state $s$ in the set $\hat S$ of (abstract) states of the policy, we define the ranking function 
$\mathit{rank}: \hat S \rightarrow \{1,\ldots,|\hat S|\}$ as the ordering of the states in $\hat S$
in the descending order according to the values $m(s)$; that is, the state with the maximal value will be
the first in the ranking. 
\end{definition}

\section{Experimental evaluation}\label{sec:evaluation}

\subsection{Research Questions}

Our goal is to demonstrate the applicability of our ranking method to
a variety of standard environments and to provide evidence of the utility
of the generated ranking. 
We aim to answer the following research questions:
\begin{description}
\item[RQ1:] How can we measure the relative importance of decisions for achieving the reward? What is a good proxy for measuring this?
\item[RQ2:] Does the approach we present in this paper scale to large policies and complex environments?
\end{description}

We answer these questions in Section~\ref{sec:prunedpol} by performing extensive experiments with 
various environments and policies. In Section~\ref{sec:discussion}, we discuss
possible applications of our ranking (including interpretability), and the effect 
of choosing a different default action.

\subsection{Experimental setup}\label{sec:setup}

We experimented in several environments. The first is \emph{Minigrid}~\cite{gym_minigrid}, 
a gridworld in which the agent operates 
with a cone of vision and can navigate many different grids,
making it more complex than a standard gridworld. In each step the agent can turn or move forward.
We also used \emph{CartPole}~\cite{DBLP:journals/corr/BrockmanCPSSTZ16}, the classic control problem 
with a continuous state-space. Finally, to test our ability to scale, we ran experiments with 
\emph{Atari games}~\cite{DBLP:journals/corr/BrockmanCPSSTZ16}.

We use policies that are trained using third-party code.
No state abstraction is applied to the gridworld environments (i.e., $\alpha$ is the identity).
The state abstraction function for the CartPole environment consists 
of rounding the components of the state vector between 0 and 2 decimal places, 
and then taking the absolute value. For 
the Atari games, as is typically done,
we crop the game's border, 
grey-scale, down-sample to $18\times14$, and lower 
the precision of pixel intensities
to make the enormous state space manageable. Note that these 
abstractions are not a contribution of ours, and were primarily chosen
for their simplicity. 
For our main experiments, we use ``repeat previous action''
as the default action. 

Examples of some environments, and important states found 
in them, are given in Figs.~\ref{fig:vis_1} and~\ref{fig:vis_2}.
Details about the state abstraction functions,
policy training, hyperparameters, etc., are provided in 
\ifArxiv{Appendix~\ref{sec:exp_setup}}.

We define two further measures in addition to the SBFL ones 
in Eq.~(\ref{eq:measures}), for comparison. Eq.~(\ref{eq:freqVis}) 
measures how frequently the state was encountered in the test suite. 
Eq.~(\ref{eq:rand}) is a random ranking
of the states visited by the test suite. We use the FreqVis measure as a
baseline because we are not aware of any previous work with the same goals 
as ours for ranking policy decisions, and a na\"ive approach to determining importance may be simply 
looking at how frequently a state is visited.

\begin{subequations}\label{eq:measures_baseline}%
\begin{tabular}{@{\hspace{-0.2cm}}l@{}l}%
\begin{minipage}[b]{6.6cm}%
\centering
\begin{equation}\text{FreqVis:}\,\,\, a_\mathit{ep}^s+a_\mathit{ef}^s+a_\mathit{np}^s+a_\mathit{nf}^s
\label{eq:freqVis}
\end{equation}
\end{minipage}
& \begin{minipage}[b]{5.8cm}%
\centering
\begin{equation}\text{Rand:}\,\,\,  \sim\text{Unif}(0, 1)
\label{eq:rand}
\end{equation}
\end{minipage}
\end{tabular}
\end{subequations}

\subsection{Experimental results}
\label{sec:prunedpol}

\paragraph{Performance of pruned policies}
The precise ranking of decisions according to their importance for the reward is intractable for all 
but very simple policies. To answer {\bf RQ1}, in our experiments, we
use the performance of 
\emph{pruned policies} as a \emph{proxy} for the quality of the ranking computed by our algorithm.
In pruned policies, the default action is used in all but the top ranked states.
For a given $r$ (a fraction or a percentage), we denote
by $\mathit{rank}[r]$ the subset of $r$ top-ranked states.
We denote by~$\pi^r$ the pruned policy obtained by \emph{pruning}
all but the top-$r$ ranked states. That is, an execution of~$\pi^r$ retains actions in the $r$ fraction
of the most important states from the original policy $\pi$ and replaces the rest
by default actions. The states that are in
$\mathit{rank}[r]$ are called the \emph{original states}.
We measure the performance of the pruned policies for increasing values of $r$ relative to
the performance of the original policy $\pi$.

These results are given in first four columns of Tab.~\ref{tab:pruning},
and some are represented graphically in Figs.~\ref{fig:graphs}a,c.
In these, SBFL stands for the \emph{SBFL portfolio}, i.e.,
the combination of the four measures in Eq.~(\ref{eq:measures}), where
the best result is taken at each point. 
The results show that the pruned policies obtained using the SBFL ranking 
can achieve performance that is comparable to $\pi$
with less than $40\%$ of the original decisions (and in 
some cases the number is as low as $20\%$).
The performance of SBFL-based pruning 
is compared with random pruning in Eq.~(\ref{eq:rand}) and
FreqVis pruning in Eq.~(\ref{eq:freqVis}). At first glance, it seems that the FreqVis pruned policies perform well in Figs.~\ref{fig:graphs}a,c.
To better understand this, we show in the last two columns of Tab.~\ref{tab:pruning} and 
in Figs.~\ref{fig:graphs}b,d how performance evolves with the proportion 
of steps in which the original policy is used over the default policy (i.e.~not replaced by the default action).
As shown in Figs.~\ref{fig:2} and~\ref{fig:6}, FreqVis does much worse by 
this metric, because it yields a pruned policy that
prefers to use the original policy as often as possible.


\begin{figure}
    \centering
    \begin{subfigure}[t]{0.45\linewidth}
        \centering
        \includegraphics[width=.8\linewidth]{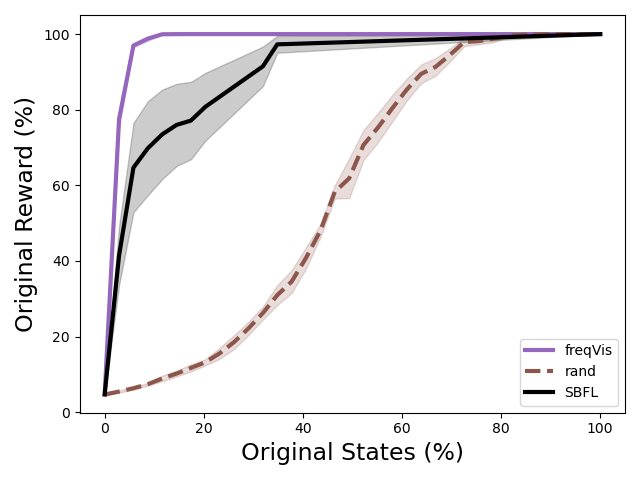}
        \caption{CartPole States in Pruned Policy}
        \label{fig:1}
    \end{subfigure}%
    \begin{subfigure}[t]{0.45\linewidth}
        \centering
        \includegraphics[width=.8\linewidth]{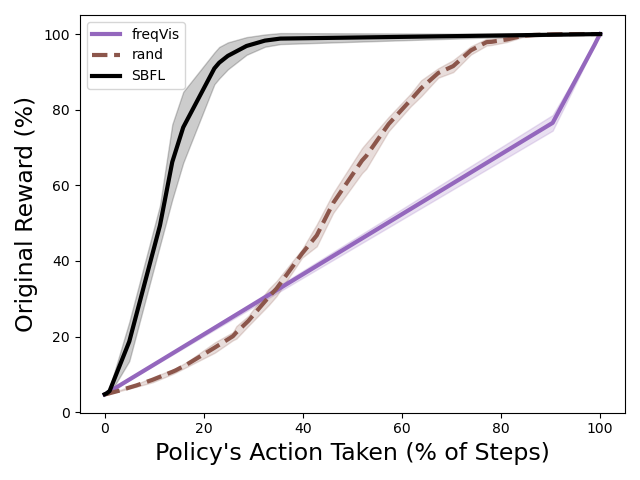}
        \caption{CartPole Actions Taken}
        \label{fig:2}
    \end{subfigure}

    \begin{subfigure}[t]{0.45\linewidth}
        \centering
        \includegraphics[width=.8\linewidth]{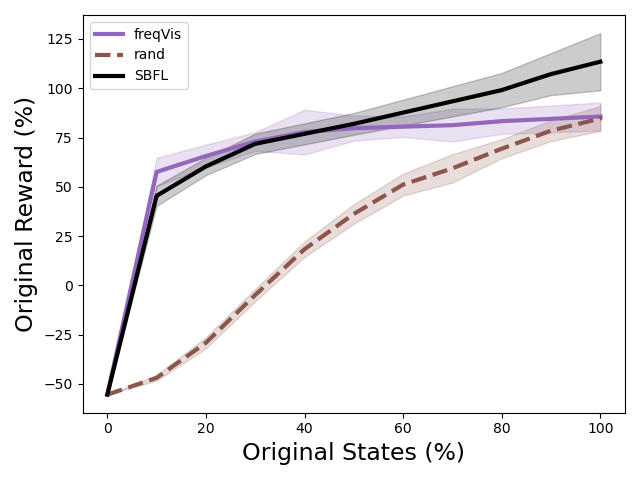}
        \caption{Krull States in Pruned Policy}
        \label{fig:5}%
    \end{subfigure}
    \begin{subfigure}[t]{0.45\linewidth}
        \centering
        \includegraphics[width=.8\linewidth]{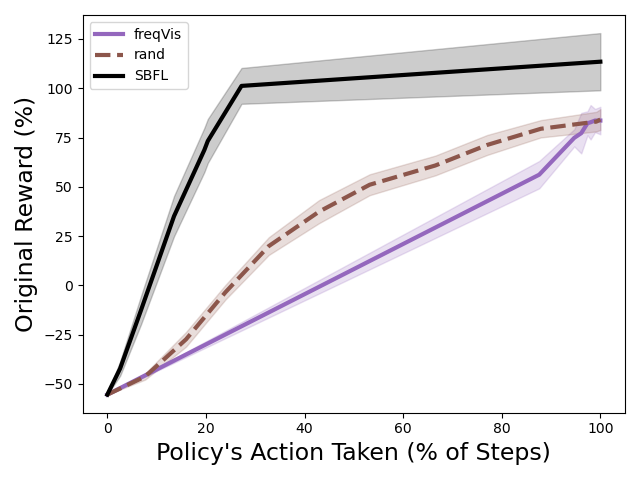}
        \caption{Krull Actions Taken}
        \label{fig:6}
    \end{subfigure}

    \caption{
    Performance of the pruned policies,
    measured as
    a percentage of the original reward.
    \textbf{(a)},\textbf{(c)}: 
    The $x$-axis is the \% of states where the original action is taken
    out of the set of all states encountered in the test suite.
    \textbf{(b)},\textbf{(d)}:
    The $x$-axis is the expected \%
    of steps in which the original policy is 
    followed over the default action during 
    an execution of the pruned policy. See the supplementary material for more detail.}
    \label{fig:graphs}
\end{figure}

\begin{table*}[t]
	\centering
    \caption{Minimum percentage of original states in pruned policies, and percentage
    of steps in which the original policy is used, before recovering 90\% 
    of original performance. Using default action ``repeat previous action''. Results are reported 
    for the SBFL portfolio ranking and the random ranking. ``\fail'' 
    denotes that the required reward was never reached,
    cf.~Sec.~\ref{sec:prunedpol}.
	}
	\label{tab:pruning}
	\newcolumntype{R}{>{\raggedleft\arraybackslash$}X<{$}}
	\newcolumntype{Y}{>{\centering\arraybackslash$}X<{$}}
	\begin{tabularx}{\linewidth}{@{}lYYYY}
		\toprule
        & \multicolumn{2}{c}{\% of original states restored} & \multicolumn{2}{c}{\% of steps that use $\pi$} \\ 
        \cmidrule{2-5}
        Environment & \mbox{SBFL} & \mbox{random} & \mbox{SBFL} & \mbox{random} \\
        \toprule
		MiniGrid & \mathbf{49} \pm 00  & 99 \pm 00 & \mathbf{76} \pm 00 & 98 \pm 01 \\
        \midrule
        Cartpole & \mathbf{31} \pm 04 & 65 \pm 02 & \mathbf{22} \pm 04 & 69 \pm 02 \\ 
        \midrule
        Alien & \fail & \fail & \fail & \fail \\
        Assault & \mathbf{45} \pm 07 & 100 \pm 00 & \mathbf{93} \pm 01 & 100 \pm 00 \\
        Atlantis & \mathbf{50} \pm 00 & 100 \pm 00 & \mathbf{99} \pm 00 & 100 \pm 00 \\
        BankHeist & \fail & \fail & \fail & \fail \\
        BattleZone & \mathbf{30} \pm 00 & 86 \pm 07 & \mathbf{84} \pm 07 & 84 \pm 08 \\
        Berzerk & \mathbf{47} \pm 12 & 100 \pm 00 & \mathbf{88} \pm 03 & 100 \pm 00 \\
        Boxing & \fail & \fail & \fail & \fail \\
        Breakout & \mathbf{10} \pm 00 & 100 \pm 00 & \mathbf{54} \pm 00 & 99 \pm 01 \\
        Breakout (abs) & \mathbf{40} \pm 00 & 85 \pm 05 & \mathbf{41} \pm 00 & 81 \pm 06 \\
        ChpperCmmnd & \fail & \fail & \fail & \fail \\
        DemonAttack & \mathbf{20} \pm 00 & 99 \pm 01 & \mathbf{98} \pm 01 & 99 \pm 02 \\
        Hero & \mathbf{48} \pm 04 & 96 \pm 03 & \mathbf{86} \pm 08 & 96 \pm 04 \\
        IceHockey & \mathbf{65} \pm 20 & \fail & \mathbf{91} \pm 08 & \fail \\
        Jamesbond & \mathbf{30} \pm 13 & 68 \pm 06 & \mathbf{59} \pm 20 & 67 \pm 06 \\
        Krull & \mathbf{75} \pm 12 & 99 \pm 01 & \mathbf{35} \pm 21 & 98 \pm 02 \\
        Phoenix & \mathbf{30} \pm 00 & 92 \pm 07 & 97 \pm 00 & \mathbf{92} \pm 07 \\
        Pong & \mathbf{21} \pm 03 & 79 \pm 03 & \mathbf{42} \pm 01 & 78 \pm 03 \\
        Qbert & \mathbf{40} \pm 00 & 100 \pm 00 & \mathbf{84} \pm 04 & 100 \pm 00 \\
        Riverraid & \mathbf{95} \pm 05 & 100 \pm 00 & \mathbf{99} \pm 01 & 100 \pm 00 \\
        RoadRunner & \fail & \fail & \fail & \fail \\
        Seaquest & \mathbf{48} \pm 04 & 94 \pm 04 & 92 \pm 04 & \mathbf{91} \pm 05 \\
        SpaceInvaders & \mathbf{30} \pm 00 & 100 \pm 00 & \mathbf{93} \pm 00 & 100 \pm 00 \\
        StarGunner & \mathbf{40} \pm 00 & 100 \pm 00 & \mathbf{99} \pm 00 & 100 \pm 00 \\
        YarsRevenge & \fail & \fail & \fail & \fail \\
		\bottomrule
    \end{tabularx}
    \normalsize
\end{table*}

We observe that using our ranking method enables a
significant pruning of the policies, while maintaining performance.
To answer {\bf RQ2}, we experimented with larger and more complex
Atari environments. The results demonstrate that our framework is
reasonably scalable, but also that the quality of the ranking is significantly
improved by using a state abstraction.
The results in the larger and more complex Atari environments show the effect 
of using a test suite that is too small: many states are not 
encountered during the computation of the ranking. This means
that $\mathit{rank}[1]$, in which the original policy is 
used in \emph{all} of the states discovered while making the ranking,
does not recover the performance of the original policy. 
Increasing the size of the test
suite would help, but this is not
a scalable solution. Instead, we use better \emph{state abstractions},
which reduce the state space.
In CartPole, this allows us to tackle a continuous domain. 
In the Atari games, even the generic abstraction we use in all the games is
sometimes not enough
(``\fail'' in Tab.~\ref{tab:pruning}).

To show the potential utility of specialised abstractions, we create one 
for \textit{Breakout} in which we extract the coordinates of the ball 
and paddle. The results obtained with this abstraction
are given in Tab.~\ref{tab:pruning} in the row labelled ``Breakout (abs)''.
While the new abstraction does worse in 
terms of states pruned within the policy, it allows us to reach 90\% 
of the policy's original performance with 13\% fewer steps in which we 
use $\pi$ over the default action.

\subsection{Discussion}
\label{sec:discussion}

\paragraph{SBFL ranking for better understanding the RL policy}
Any strong claim about the ranking's application to interpretability requires a user study, which is out of scope of this paper.
However, we can look to existing research looking at the usefulness of SBFL. While some studies suggest that the users typically do not go over the list of possible
causes generated by SBFL linearly (and hence question the usefulness of the ranking)~\cite{PO11}, a recent
large-scale study demonstrates statistically significant and substantial improvements for the users
who use an SBFL tool, and the results hold even for ``mediocre'' SBFL tools~\cite{XBLL16}.
Based on this evidence,
we suggest that the ranking can be used to explain policy decisions,
as the ranked list itself would be helpful to identify the most important decisions.
In addition, the pruned policies that we construct
are simpler than the original policies while achieving a comparable performance, which can make identifying problems 
more straightforward. Examples are 
presented in the CartPole and Minigrid
sections of the website. Finally, in Fig.~\ref{fig:heatmap}, 
we show a heatmap of the scores of each state of a minigrid environment. 
In each grid square, the agent can be facing four directions. 
We show the score based on the tarantula measure
for these directions from blue (lowest) to red (highest). We show this only for
the states visited along a path to the goal.
Our heatmap gives information 
about the general behaviour of the policy. In this case, points at which 
the agent needs to turn are more red (more important) than points where 
the agent is walking in a straight line. To understand why the downward-facing
states in the right-most column are considered important, refer to our 
website for a full heatmap and explanation.

\begin{figure}
    \centering
    \includegraphics[width=0.25\linewidth]{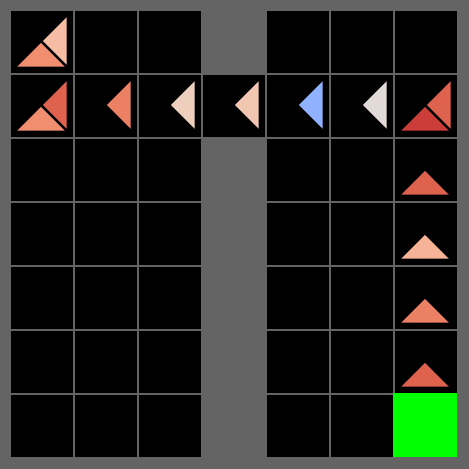}
    \caption{Example of a heatmap made based on scores.
        Colour from least to most important are blue, white and red.
        \textbf{(a)}
        Our heatmap based on the Tarantula measure, 
        showing all the states an agent encounters while 
        walking to the goal, including the direction in which 
        the agent was facing. For example, in the top left grid location,
        we show the importance of the state in which the agent is facing 
        right, and the state in which it is facing down.}
    \label{fig:heatmap}
\end{figure}

\paragraph{A Different Default Action}
Not all environments have an obvious default action.
In this case, a straightforward choice is to set the default action to ``take a random action''. 
We measured the effect of this choice by running the same experiments with the changed
default action, and detailed results are available in \ifArxiv{Appendix~\ref{sec:randact}}. 
The results are similar or slightly worse to the ones obtained with the default action ``repeat the previous action''.
In most games, the difference was small ($\leq 10\%$); a few games show a marginal improvement.
These results indicate that the choice of a default action has no effect on our conclusions. 


\paragraph{Good vs.~bad policies}
Finally, we performed some initial work towards using the ranking to understand the 
difference between high and low-performing policies. To this end, we produced
a ranking according to the high-performing policy, and then compared how the two
policies behaved in the highest ranked states. Interestingly, our experiments demonstrate that
the high performing and low performing policies agree on $80\%$ of the actions in the top $10\%$ of states. 
This suggests that the policy training (in CartPole)
first picks the
`low-hanging fruit' by performing well in the most
important states. The difference between a high-performing and a low-performing 
policy is mostly in the 
lower-ranked states. Full results for this experiment are 
available in \ifArxiv{Fig.~\ref{fig:cpolegoodbad} in the Appendix}.

\section{Related work}
\label{sec:related}


There is a significant body of work on identifying the important parts of
trained algorithms, but to the best of our knowledge none suggested to rank
the states as we do.  Prioritised experience
replay~\cite{DBLP:journals/corr/SchaulQAS15} looks for the most important
\emph{transitions} for training.  Saliency
maps~\cite{greydanus2017visualizing,wang2015dueling,zahavy2016graying}
identify the parts of the state that most influence influence
the agent’s decision. \citet{schk2020} have previously applied SBFL  to visual feature
importance for input images given to image classifiers.
Other attempts include identifying the important parts of the \textit{representation} of
the policy by looking at the parameters of a trained model and pruning it to reduce
its size~\cite{livne2020pops}.  None of these methods attempt to understand
what the important decisions of the policy as a whole are.

Much of the recent work focuses on making deep learning models more
interpretable~\cite{lime,shap,schk2020}.  Many
approaches~\cite{greydanus2017visualizing,wang2015dueling,zahavy2016graying}
to explaining deep reinforcement learning methods explain the decision made in a single
state, without the context of the past or the future behaviour. 
\citet{iyer2018transparency} explain a single decision via an object-level
saliency map by leveraging the pixel-level explanation and object
detection. 
As these methods focus on single decisions, the explanation is
typically not sufficient to understand the overall decision-making of the trained agent.

Other work has also attempted to explain entire policies, rather than individual decisions. 
\citet{ehsan2018rationalization} produce natural language explanations for state-action pairs,
based on a human-provided corpus of explanations. \citet{topin2019generation} 
create a Markov chain which acts as an 
abstraction of the policy, making it easier to reason about the policy. They 
create the Markov chain by grouping states into abstract states based 
on how similarly the policy acts in those states; our method is 
substantially different in that it ranks states based on importance, 
rather than grouping similar states. While our method allows for the 
use of abstract states, it is not always necessary, and we are not 
contributing any specific abstraction function.
Similarly, \citet{DBLP:conf/aips/SreedharanSK20} propose a 
method for creating a temporal abstraction of a policy by using 
bottleneck `landmarks' in the policy's executions. A robot's behaviour
can be explained using operator-specified
``important program state variables'' and ``important functions'' \cite{hayes2017improving}.  
We find that the policy-wide decision ranking in this paper is an easier and more
general method for understanding the policy.
For more work in this vein, we encourage the reader to consult the overview from
\citet{DBLP:conf/ijcai/ChakrabortiSK20} of the 
rapidly growing field of Explainable AI Planning (XAIP).

There have been attempts to make more interpretable models, either from
scratch~\cite{hein2018interpretable}, or by approximating a trained neural
network~\cite{verma2018programmatically}.  In the latter case, our method
may be useful for determining in which states the approximation must be most
accurate.

\section{Conclusions}

We have applied SBFL-based ranking of states to
reinforcement learning and demonstrated that this ranking correlates
with the relative importance of states for the policy's performance.
The ranking can be used to explain RL policies, similarly to the way ranked
program locations are used to understand the causes of a bug.  
We evaluate the quality of our ranking by constructing simpler
pruned policies, where only the most important decisions are made
according to the policy, and the rest are default. Our experiments show that the performance of the pruned policies is comparable to the performance of the original policies,
thus supporting our claim that the SBFL-based ranking is accurate.
Moreover, the pruned policies may be preferable in many use-cases, as they are simpler.
Our approach can be scaled with the use of abstractions, as demonstrated
by the larger Atari environments. 

In the future, we hope to explore different applications 
of the ranking, new measures, more nuanced hyperparameter selection,
and relaxing the binary constraint over the assertion $C$.
We do not expect our work to have any negative societal impacts, as it only serves to improve our understanding of the policies we train; the main
concern is incorrectly increasing confidence in a policy.

\noindent \paragraph{\textbf{Funding transparency statement}}
The authors acknowledge funding from the UKRI Trust-worthy Autonomous Systems Hub (EP/V00784X/1) and the UKRI Strategic Priorities Fund to the UKRI Research Node on Trustworthy Autonomous Systems Governance and Regulation (EP/V026607/1).

\bibliographystyle{plainnat}
\bibliography{paper}  

\iftoggle{arxiv}{\clearpage
\appendix

\section{Experiment Setup}
\label{sec:exp_setup}

Code for experiments can be found 
at the github repository: 
\url{https://github.com/hadrien-pouget/Ranking-Policy-Decisions}.

\subsection{Environments}

\paragraph{MiniGrid}
We use the SimpleCrossingS9N1-v0 variant of the Minimalistic 
Gridworld Environment (MiniGrid)~\cite{gym_minigrid}. 
The environment always
consists of a $7\times7$ grid in which the agent can 
navigate, with a wall separating the agent's starting 
position from the goal. There is always on hole in the wall.
At each episode, the position of the wall and hole are randomised.
The agent is rewarded for reaching the goal, and this reward
is linearly annealed from 1 to 0.1 by the final step. After 322 steps, the episode
ends. The state information given the agent is a cone of vision,
allowing a single agent to 

\paragraph{CartPole}
We use OpenAI's `Gym'~\cite{DBLP:journals/corr/BrockmanCPSSTZ16} 
implementation of the CartPole environment. The agent may 
push the cart left or right to balance the pole, and 
gains a reward of 1 at each step. If the pole falls too 
far to the side, or the cart moves too far to the side, 
the episode is terminated. The episode ends after 200 steps.

\paragraph{Atari Games}
To implement the Atari games, we use OpenAI's `Gym'~\cite{DBLP:journals/corr/BrockmanCPSSTZ16}
package.
We artificially limit the length of each episode to 
600 steps to reduce running time of analyses.
For all the games, we use the 
NoFrameskip-V4 versions.

\subsection{Abstractions}
\label{sec:abstractions}

\begin{figure}[t]
	\centering
	\begin{subfigure}[t]{0.48\linewidth}
		\includegraphics[width=\linewidth]{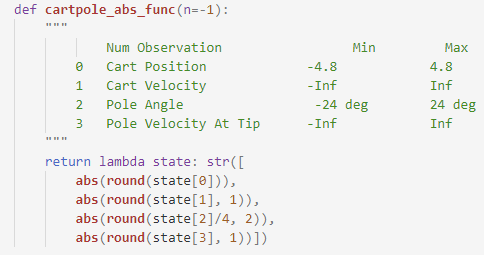}
		\caption{CartPole abstraction function}
	\end{subfigure}\hfill%
	\begin{subfigure}[t]{0.48\linewidth}
		\includegraphics[width=\linewidth]{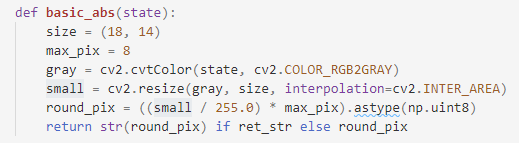}
		\caption{Atari Games generic abstraction function}
	\end{subfigure}
	\caption{Abstraction Functions in Python. ``abs'' is the 
	absolute value, and ``round'' rounds the input to the 
	number of decimal points given.}
	\label{fig:abstractions}
\end{figure}

\begin{figure}[t]
	\centering
	\begin{subfigure}[t]{0.24\linewidth}
	    \centering
		\includegraphics[width=\linewidth]{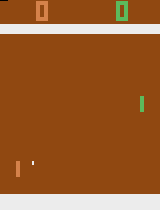}
		\caption{Pong before abstraction}
	\end{subfigure}\hfill%
	\begin{subfigure}[t]{0.24\linewidth}
	    \centering
		\includegraphics[width=\linewidth]{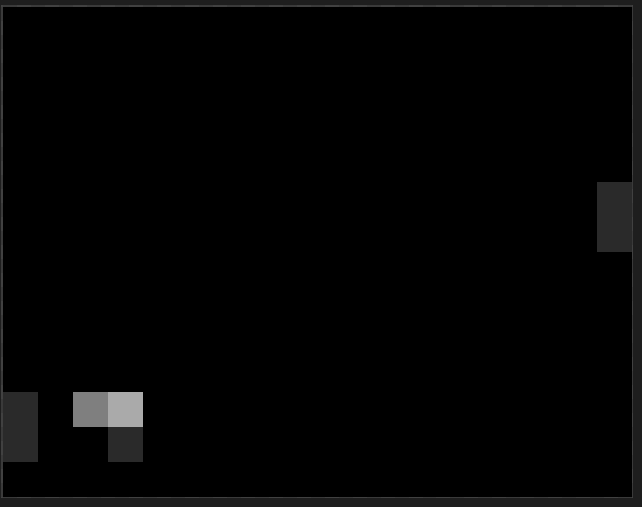}
		\caption{Pong after abstraction}
	\end{subfigure}\hfill%
	\begin{subfigure}[t]{0.24\linewidth}
	    \centering
		\includegraphics[width=\linewidth]{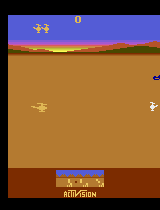}
		\caption{Chopper Command before abstraction}
	\end{subfigure}\hfill%
	\begin{subfigure}[t]{0.24\linewidth}
	    \centering
		\includegraphics[width=\linewidth]{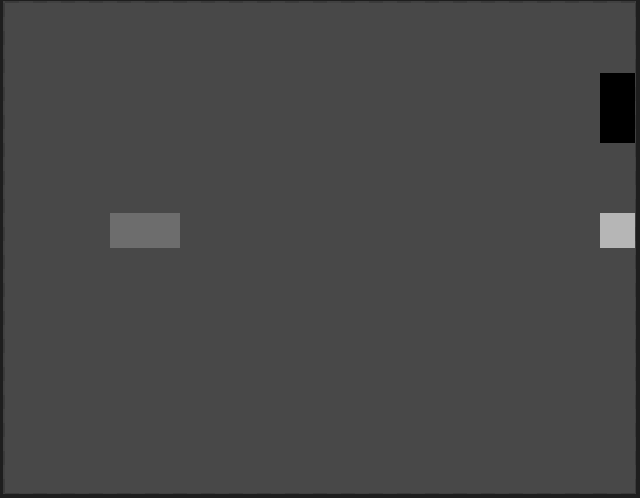}
		\caption{Chopper Command after abstraction}
	\end{subfigure}
\caption{State abstractions for two Atari games}
\label{fig:atariabs}
\end{figure}

The functions used for abstracting the state
space are shown in 
Figure~\ref{fig:abstractions}. 

\paragraph{MiniGrid} No abstraction function is used 
in Minigrid experiments.

\paragraph{CartPole} The state space is 
continuous in CartPole, and we discretize it by rounding
each element of the state. The pole angle 
is first divided by 4, to further reduce the space.
We finally take the absolute value to exploit CartPole's symmetry.
It is worth noting that this would not work with any default
action. For example, if the default action were 'move right',
then states where the pole is leaning left (and moving right would 
make it fall over) would be more important than states where the 
pole is leaning right (and moving right would straighten it).
We repeat the last action as our default action, and this 
treats both side symmetrically.

\paragraph{Atari Games}
For Atari games, the state is 
cropped to only the game area,
and important parts of the state (e.g. character) 
are made bigger (so that they appear even in the abstract
state) before applying the abstraction function. The 
abstraction function grey-scales, downsamples 
to $18\times14$ pixels, and then forces the pixels 
to take integer values between 0 and 8.
Examples of the Atari games abstract states are shown in 
Figure~\ref{fig:atariabs}.

In the specialised Breakout abstraction, 
the pixel coordinates of the ball and paddle 
are isolated and rounded to the nearest 10.

\subsection{Policies}

\paragraph{Minigrid} For MiniGrid, the policy was trained using code from the 
``RL Starter Files'' github repository~\cite{willems_lcswillemsrl-starter-files_2020},
and achieves an expected reward of 0.88.

\paragraph{CartPole} For CartPole, the policy was trained using code from the 
``PyTorch CartPole Example'' github repository~\cite{kang_g6lingreinforcement-learning-pytorch-cartpole_2020},
with expected reward 200.
We include a low-performing policy in our experiment, with expected reward 97.
We have obtained this policy by stopping the training early.

\paragraph{Atari Games}
The policies we use for the Atari games are taken from Uber's Atari Zoo
package~\cite{DBLP:conf/ijcai/SuchMLWCLZSBCL19}. 
The performance for each game is given in Table~\ref{tab:ataripol}.

\begin{table}[t]
    \centering
    \caption{Expected reward for trained policy and
    random behaviour in each Atari game, constrained
    to 600 steps.}
    \begin{tabular}{lrr}
        \toprule
        Game & Policy Reward & Random Reward \\
        \midrule
        Alien & 1500.0 & 178.0 \\
        Assault & 929.1 & 214.2 \\
        Asteroids & 301.0 & 564.0 \\
        Atlantis & 14170.0 & 7950.0 \\
        BankHeist & 553.0 & 12.0 \\
        BattleZone & 12300.0 & 1400.0 \\
        Berzerk & 811.0 & 140.0 \\
        Boxing & 97.8 & -0.1 \\
        Breakout & 19.6 & 1.4 \\
        ChopperCommand & 3510.0 & 600.0 \\
        DemonAttack & 133.5 & 91.0 \\
        Hero & 13414.0 & 272.0 \\
        IceHockey & 0.2 & -2.0 \\
        Jamesbond & 290.0 & 15.0 \\
        Krull & 1159.1 & 444.0 \\
        Phoenix & 1766.0 & 390.0 \\
        Pong & 6.7 & -13.1 \\
        Qbert & 7620.0 & 102.5 \\
        Riverraid & 5485.0 & 1268.0 \\
        RoadRunner & 30020.0 & 20.0 \\
        Seaquest & 598.0 & 74.0 \\
        SpaceInvaders & 724.5 & 164.5 \\
        StarGunner & 10000.0 & 330.0 \\
        Tennis & -1.0 & -8.3 \\
        YarsRevenge & 21008.6 & 2304.6 \\
        \bottomrule
    \end{tabular}
    \label{tab:ataripol}
\end{table}

\subsection{Hyperparameters}

The default action was ``repeat the previous action'' or ``take a random action''
for all experiments. Table~\ref{tab:hyperp} shows all the 
other hyperparameters for the experiments. We also present 
$n_{test}$, the number of episodes over which each 
pruned policy is evaluated when performing experiments with 
pruned policies.

In general, setting the hyperparameters will depend on your goals. 
If you are testing a car, you might be interested in the condition $C$
``was there a crash?'', and the default action ``continue driving forwards''. 
In this case, the ranking would show you which of the policy's decisions
had the most influence on crashing, if the alternative were to continue straight ahead.
In situation in which there is no obvious default action, ``take random action'' can be used.
If there is no obvious condition, then it is general best to choose a combination of $\mu$ and $C$ which 
balances the test suite between successes and failures. In our experiments
$|\mathcal{T}(\pi)|$ and $n_{test}$ were chosen to be as high 
as possible within reasonable running times. 

\paragraph{Minigrid and Atari Games}
In Minigrid and Atari games,
$\mu = 0.2$ was empirically
found to be a good balance between mutating enough states 
to gain information, and not mutating so much as to destroy 
the policy. $C$ was chosen to balance the test suite 
between success and failure. 

\paragraph{CartPole} In CartPole, $C=$ reward $\geq$ 200 was chosen 
as it corresponds to checking whether the pole ever fell, 
or the cart went out of bounds. $\mu$ was set to balance the test suite.

\begin{table}[H]
    \centering
    \caption{Hyperparameters for each environment.}
    \begin{tabular}{lcclc}
        \toprule
        Environment & $|\mathcal{T}(\pi)|$ & $\mu$ & $C$ & $n_{test}$ \\
        \midrule
        MiniGrid & 5000 & 0.2 & reward $\geq$ 0.8 & 100 \\
        \midrule
        CartPole & 5000 & 0.4 & reward $\geq$ 200 & 100 \\
        \midrule
        Alien & 1000 & 0.2 & reward $\geq$ 90.0 & 50 \\
        Assault & 1000 & 0.2 & reward $\geq$ 1.0 & 50 \\
        Atlantis & 1000 & 0.2 & reward $\geq$ 400.0 & 50 \\
        BankHeist & 1000 & 0.2 & reward $\geq$ 0.0 & 50 \\
        BattleZone & 1000 & 0.2 & reward $\geq$ 0.0 & 50 \\
        Berzerk & 1000 & 0.2 & reward $\geq$ 0.0 & 50 \\
        Boxing & 1000 & 0.2 & reward $\geq$ 0.0 & 50 \\
        Breakout & 1000 & 0.2 & reward $\geq$ 0.0 & 50 \\
        ChopperCommand & 1000 & 0.2 & reward $\geq$ 0.0 & 50 \\
        DemonAttack & 1000 & 0.2 & reward $\geq$ 0.0 & 50 \\
        Hero & 1000 & 0.2 & reward $\geq$ 910.0 & 50 \\
        IceHockey & 1000 & 0.2 & reward $\geq$ 0.0 & 50 \\
        Jamesbond & 1000 & 0.2 & reward $\geq$ 0.0 & 50 \\
        Krull & 1000 & 0.2 & reward $\geq$ 21.5 & 50 \\
        Phoenix & 1000 & 0.2 & reward $\geq$ 0.0 & 50 \\
        Pong & 1000 & 0.2 & reward $\geq$ 0.0 & 50 \\
        Qbert & 1000 & 0.2 & reward $\geq$ 75.0 & 50 \\
        Riverraid & 1000 & 0.2 & reward $\geq$ 80.0 & 50 \\
        RoadRunner & 1000 & 0.2 & reward $\geq$ 600.0 & 50 \\
        Seaquest & 1000 & 0.2 & reward $\geq$ 0.0 & 50 \\
        SpaceInvaders & 1000 & 0.2 & reward $\geq$ 0.0 & 50 \\
        StarGunner & 1000 & 0.2 & reward $\geq$ 0.0 & 50 \\
        YarsRevenge & 1000 & 0.2 & reward $\geq$ 176.0 & 50 \\
                \bottomrule
        \end{tabular}
        \label{tab:hyperp}
\end{table}

\subsection{Computational Resources}

Our method does not require any re-training, an often expensive step.
The primary cost is performing executions of the agent in the environment. 
To calculate the rankings, this means doing as many executions as are in the test suite. 
In our experiments, this was 500 for Atari games, which is far less than most 
training algorithms.
The time taken to compute each rankings for each of the environments in our main experiments is given in Table~\ref{tab:time}.
For all of our experiments, we used an Intel(R) Xeon(R) Silver 
4114 CPU with 2.20\,GHz, and 188\,GB RAM, and a 16\,GB Tesla V100 GPU.

Producing the test suite takes time linear in the size of the test suite,
and computing the ranking require sorting the set of all of the visited states $S$, which 
is $O(|S|\log|S|)$.

\begin{table}[H]
    \centering
    \caption{Average time to compute ranking in each environment.}
    \begin{tabular}{lc}
        \toprule
        Environment & Time (mm.ss) \\
        \midrule       
        MiniGrid & 28.40 \\
        \midrule
        CartPole & 09.40 \\
        \midrule
        Alien & 34.51 \\
        Assault & 34.41 \\
        Atlantis & 32.50 \\
        BankHeist & 31.54 \\
        BattleZone & 34.52 \\
        Berzerk & 21.35 \\
        Boxing & 35.37 \\
        Breakout & 31.19 \\
        ChopperCommand & 33.08 \\
        DemonAttack & 31.11 \\
        Hero & 35.08 \\
        IceHockey & 34.27 \\
        Jamesbond & 32.45 \\
        Krull & 28.20 \\
        Phoenix & 31.14 \\
        Pong & 31.41 \\
        Qbert & 29.48 \\
        Riverraid & 33.46 \\
        RoadRunner & 34.37 \\
        Seaquest & 31.18 \\
        SpaceInvaders & 31.29 \\
        StarGunner & 32.03 \\
        YarsRevenge & 31.04 \\
        \bottomrule
        \end{tabular}
    \label{tab:time}
\end{table}

\newpage
\section{Results}

Additional results, including heatmaps and 
graphs showing each measure individually,
can be found at 
\url{https://www.cprover.org/deepcover/neurips2021/}.

\subsection{More Results from Policy Pruning}
Table~\ref{tab:pruning_50} is complementary to Table~\ref{tab:pruning},
and shows how many states were restored, or in how many states 
the policy was used over the default action, in order to reach 
50\% of the original policy's performance.

\begin{table*}[t]
	\centering
    \caption{Minimum percentage of original states in pruned policies, and percentage
    of steps in which the original policy is used, before recovering 50\% 
    of original performance. Using default action ``repeat previous action''. Results are reported 
    for the SBFL portfolio ranking and the random ranking. ``\fail'' 
    denotes that the required reward was never reached,
    cf.~Sec.~\ref{sec:prunedpol}.
	}
	\label{tab:pruning_50}
	\newcolumntype{R}{>{\raggedleft\arraybackslash$}X<{$}}
	\newcolumntype{Y}{>{\centering\arraybackslash$}X<{$}}
	\begin{tabularx}{\linewidth}{@{}lYYYY}
		\toprule
        & \multicolumn{2}{c}{\% of original states restored} & \multicolumn{2}{c}{\% of steps that use $\pi$} \\ 
        \cmidrule{2-5}
        Environment & \mbox{SBFL} & \mbox{random} & \mbox{SBFL} & \mbox{random} \\
        \toprule
        MiniGrid & \textbf{35} \pm  00 & 85 \pm 00 & \textbf{49} \pm 00 & 66 \pm 04) \\
        \midrule
        Cartpole &  \textbf{6} \pm 04 & 46 \pm 01 & \textbf{13} \pm 01 & 47 \pm 01 \\ 
        \midrule
        Alien & \fail & \fail & \fail & \fail\\
        Assault & \mathbf{22} \pm 04 & 54 \pm 05 & 57 \pm 10 & \mathbf{51} \pm 05\\
        Atlantis & \mathbf{20} \pm 00 & 90 \pm 00 & \mathbf{87} \pm 00 & 92 \pm 01\\
        BankHeist & \mathbf{35} \pm 05 & 91 \pm 03 & \mathbf{62} \pm 03 & 83 \pm 06\\
        BattleZone & \mathbf{30} \pm 00 & 52 \pm 04 & 76 \pm 01 & \mathbf{42} \pm 05\\
        Berzerk & \mathbf{10} \pm 00 & 88 \pm 03 & \mathbf{74} \pm 10 & 80 \pm 06\\
        Boxing & \fail & \fail & \fail & \fail\\
        Breakout & \mathbf{10} \pm 00 & 82 \pm 04 & \mathbf{54} \pm 00 & 67 \pm 07\\
        Breakout (abs) & \mathbf{40} \pm 00 & 61 \pm 03 & \mathbf{41} \pm 00 & 49 \pm 05\\
        ChpperCmmnd & \mathbf{23} \pm 05 & 83 \pm 06 & \mathbf{80} \pm 05 & 80 \pm 10\\
        DemonAttack & \mathbf{20} \pm 00 & 90 \pm 05 & 95 \pm 06 & \mathbf{84} \pm 08\\
        Hero & \mathbf{32} \pm 05 & 68 \pm 04 & \mathbf{45} \pm 06 & 58 \pm 06\\
        IceHockey & \mathbf{20} \pm 04 & 74 \pm 05 & \mathbf{66} \pm 04 & 67 \pm 05\\
        Jamesbond & \mathbf{20} \pm 00 & 40 \pm 00 & \mathbf{28} \pm 07 & 38 \pm 01\\
        Krull & \mathbf{19} \pm 03 & 64 \pm 05 & \mathbf{20} \pm 05 & 57 \pm 05\\
        Phoenix & \mathbf{20} \pm 00 & 78 \pm 06 & \mathbf{50} \pm 00 & 74 \pm 07\\
        Pong & \mathbf{10} \pm 00 & 50 \pm 00 & \mathbf{33} \pm 02 & 42 \pm 01\\
        Qbert & \mathbf{30} \pm 00 & 89 \pm 03 & \mathbf{78} \pm 01 & 78 \pm 04\\
        Riverraid & \mathbf{38} \pm 04 & 85 \pm 05 & 82 \pm 02 & \mathbf{80} \pm 06\\
        RoadRunner & \mathbf{28} \pm 04 & 85 \pm 05 & \mathbf{77} \pm 04 & 79 \pm 07\\
        Seaquest & \mathbf{30} \pm 01 & 70 \pm 01 & \mathbf{55} \pm 09 & 56 \pm 02\\
        SpaceInvaders & \mathbf{22} \pm 04 & 81 \pm 03 & \mathbf{63} \pm 15 & 75 \pm 04\\
        StarGunner & \mathbf{20} \pm 00 & 70 \pm 00 & 71 \pm 13 & \mathbf{69} \pm 02\\
        YarsRevenge & \mathbf{27} \pm 04 & 82 \pm 04 & \mathbf{65} \pm 04 & 70 \pm 06\\
		\bottomrule
    \end{tabularx}
    \normalsize
\end{table*}

\subsection{Results with Random Action as Default Action}
\label{sec:randact}
We repeat the experiments from Tables~\ref{tab:pruning}\&\ref{tab:pruning_50}
with the same hyperparameters, using the default action ``take a random action'',
rather than ``repeat previous action''. 
These results are shown in Tables~\ref{tab:random_pruning_90}\&\ref{tab:random_pruning_50}.

Table~\ref{tab:comparisontable} shows how changing the default action from 
``repeat previous action'' to ''take random action'' affected the results.
Red numbers show when the change worsened results, and green when it improved them.
In general, this change worsened results. However, in many cases there was little change,
or positive change. This generally reflects that in most games ``repeat previous action'' 
is a sensible default action, but that in some it is easier to replace actions with random 
actions.

\begin{table*}[h]
	\centering
    \caption{Minimum percentage of original states in pruned policies, and percentage
    of steps in which the original policy is used, before recovering 90\% 
    of original performance. Using default action ``take random action''. Results are reported 
    for the SBFL portfolio ranking and the random ranking. ``\fail'' 
    denotes that the required reward was never reached,
    cf.~Sec.~\ref{sec:prunedpol}.
	}
	\label{tab:random_pruning_90}
	\newcolumntype{R}{>{\raggedleft\arraybackslash$}X<{$}}
	\newcolumntype{Y}{>{\centering\arraybackslash$}X<{$}}
	\begin{tabularx}{\linewidth}{@{}lYYYY}
		\toprule
        & \multicolumn{2}{c}{\% of original states restored} & \multicolumn{2}{c}{\% of steps that use $\pi$} \\ 
        \cmidrule{2-5}
        Environment & \mbox{SBFL} & \mbox{random} & \mbox{SBFL} & \mbox{random} \\
        \toprule
        MiniGrid & \textbf{51} \pm  00 & 91 \pm 03 & \textbf{72} \pm 00 & 85 \pm 03 \\
        \midrule
        Cartpole &  \textbf{41} \pm 01 & 68 \pm 01 & \textbf{61} \pm 03 & 72 \pm 01 \\ 
        \midrule
        Alien & \fail & \fail & \fail & \fail\\
        Assault & \mathbf{71} \pm 07 & 100 \pm 00 & \mathbf{98} \pm 01 & 100 \pm 00\\
        Atlantis & \mathbf{40} \pm 00 & 100 \pm 00 & \mathbf{99} \pm 00 & 100 \pm 00\\
        BankHeist & \fail & \fail & \fail & \fail\\
        BattleZone & \mathbf{40} \pm 00 & 100 \pm 00 & \mathbf{98} \pm 00 & 100 \pm 00\\
        Berzerk & \fail & \fail & \fail & \fail\\
        Breakout & \mathbf{37} \pm 05 & 99 \pm 01 & \mathbf{95} \pm 01 & 99 \pm 02\\
        ChpperCmmnd & \fail & \fail & \fail & \fail\\
        DemonAttack & \mathbf{19} \pm 01 & 90 \pm 11 & 99 \pm 00 & \mathbf{89} \pm 11\\
        Hero & \mathbf{49} \pm 01 & 97 \pm 02 & \mathbf{96} \pm 01 & 97 \pm 02\\
        Phoenix & \mathbf{20} \pm 00 & 79 \pm 05 & \mathbf{71} \pm 00 & 79 \pm 06\\
        Pong & \mathbf{40} \pm 00 & 90 \pm 00 & 96 \pm 00 & \mathbf{89} \pm 00\\
        Qbert & \fail & \fail & \fail & \fail\\
        Riverraid & \fail & \fail & \fail & \fail\\
        RoadRunner & \fail & \fail & \fail & \fail\\
        SpaceInvaders & \mathbf{51} \pm 03 & 100 \pm 00 & \mathbf{95} \pm 00 & 100 \pm 00\\
        StarGunner & \mathbf{40} \pm 00 & 99 \pm 01 & \mathbf{98} \pm 00 & 99 \pm 01\\
		\bottomrule
    \end{tabularx}
    \normalsize
\end{table*}

\begin{table*}
	\centering
    \caption{Minimum percentage of original states in pruned policies, and percentage
    of steps in which the original policy is used, before recovering 50\% 
    of original performance. Using default action ``take random action''. Results are reported 
    for the SBFL portfolio ranking and the random ranking. ``\fail'' 
    denotes that the required reward was never reached,
    cf.~Sec.~\ref{sec:prunedpol}.
	}
	\label{tab:random_pruning_50}
	\newcolumntype{R}{>{\raggedleft\arraybackslash$}X<{$}}
	\newcolumntype{Y}{>{\centering\arraybackslash$}X<{$}}
	\begin{tabularx}{\linewidth}{@{}lYYYY}
		\toprule
        & \multicolumn{2}{c}{\% of original states restored} & \multicolumn{2}{c}{\% of steps that use $\pi$} \\ 
        \cmidrule{2-5}
        Environment & \mbox{SBFL} & \mbox{random} & \mbox{SBFL} & \mbox{random} \\
        \toprule
        MiniGrid & \textbf{32} \pm  00 & 61 \pm 03 & \textbf{25} \pm 00 & 48 \pm 04 \\
        \midrule
        Cartpole &  \textbf{37} \pm 01 & 48 \pm 01 & \textbf{39} \pm 00 & 49 \pm 02 \\ 
        \midrule
        Alien & \fail & \fail & \fail & \fail\\
        Assault & \mathbf{30} \pm 00 & 70 \pm 00 & 85 \pm 05 & \mathbf{68} \pm 00\\
        Atlantis & \mathbf{20} \pm 00 & 89 \pm 05 & \mathbf{89} \pm 00 & 91 \pm 05\\
        BankHeist & \mathbf{30} \pm 00 & 90 \pm 00 & \mathbf{79} \pm 00 & 98 \pm 01\\
        BattleZone & \mathbf{30} \pm 00 & 59 \pm 03 & 59 \pm 01 & \mathbf{53} \pm 03\\
        Berzerk & \mathbf{16} \pm 08 & 99 \pm 01 & \mathbf{85} \pm 07 & 98 \pm 02\\
        Breakout & \mathbf{10} \pm 00 & 71 \pm 04 & \mathbf{65} \pm 01 & 70 \pm 03\\
        ChpperCmmnd & \mathbf{83} \pm 16 & 100 \pm 00 & \mathbf{95} \pm 02 & 99 \pm 01\\
        DemonAttack & \mathbf{16} \pm 05 & 54 \pm 07 & 88 \pm 13 & \mathbf{53} \pm 06\\
        Hero & \mathbf{37} \pm 04 & 76 \pm 03 & \mathbf{63} \pm 13 & 71 \pm 04\\
        Phoenix & \mathbf{10} \pm 00 & 59 \pm 05 & \mathbf{45} \pm 00 & 56 \pm 05\\
        Pong & \mathbf{30} \pm 00 & 69 \pm 03 & \mathbf{64} \pm 00 & 65 \pm 03\\
        Qbert & \mathbf{30} \pm 00 & 100 \pm 00 & \mathbf{78} \pm 01 & 99 \pm 01\\
        Riverraid & \mathbf{40} \pm 00 & 90 \pm 00 & \mathbf{75} \pm 05 & 88 \pm 01\\
        RoadRunner & \fail & \fail & \fail & \fail\\
        SpaceInvaders & \mathbf{29} \pm 03 & 90 \pm 00 & \mathbf{86} \pm 08 & 86 \pm 01\\
        StarGunner & \mathbf{20} \pm 00 & 73 \pm 05 & 80 \pm 03 & \mathbf{73} \pm 04\\
		\bottomrule
    \end{tabularx}
    \normalsize
\end{table*}

\begin{table*}
	\centering
    \caption{How the success of the pruned policy changes 
    when switching from ``repeat previous action'' to ``take random action''
    as a default action. Results being compared are from Tables~\ref{tab:pruning},\ref{tab:pruning_50},\ref{tab:random_pruning_90},\ref{tab:random_pruning_50}. 
    Positive numbers reflect that by using the random default action, we need to use the original policy 
    in more steps before recovering either 50\% or 90\% of the original policy's performance. ``Failed'' 
    indicates that 90\% (or 50\%) of the original policy's performance was never recovered with the 
    random default action, but that it was recovered with ``repeat previous action''. ``-'' indicates 
    that it was recovered by neither; cf.~Sec.~\ref{sec:prunedpol} for more details about why these failures
    may occur.
	}
	\label{tab:comparisontable}
	\newcolumntype{R}{>{\raggedleft\arraybackslash$}X<{$}}
	\newcolumntype{Y}{>{\centering\arraybackslash$}X<{$}}
	\begin{tabularx}{\linewidth}{@{}lYYYY}
		\toprule
        & \multicolumn{2}{c}{difference in \% of original states restored} & \multicolumn{2}{c}{difference in \% of steps that use $\pi$} \\ 
        \cmidrule{2-5}
        Environment & \mbox{90\%} & \mbox{50\%} & \mbox{90\%} & \mbox{50\%} \\
        \toprule
        MiniGrid & \textcolor{red}{2} & \textcolor{Green}{-3} & \textcolor{Green}{-4} & \textcolor{Green}{-24}\\
        \midrule
        Cartpole &  \textcolor{red}{10} & \textcolor{red}{31} & \textcolor{red}{39} & \textcolor{red}{26}\\
        \midrule
        Alien & - & - & - & -\\
        Assault & \textcolor{red}{26} & \textcolor{red}{8} & \textcolor{red}{5} & \textcolor{red}{8}\\
        Atlantis & \textcolor{Green}{-10} & \textcolor{black}{0} & \textcolor{black}{0} & \textcolor{black}{0}\\
        BankHeist & - & \textcolor{Green}{-5} & - & \textcolor{Green}{-5}\\
        BattleZone & \textcolor{red}{10} & \textcolor{black}{0} & \textcolor{red}{14} & \textcolor{black}{0}\\
        Berzerk & \textcolor{red}{Failed} & \textcolor{red}{6} & \textcolor{red}{Failed} & \textcolor{red}{6}\\
        Breakout & \textcolor{red}{27} & \textcolor{black}{0} & \textcolor{red}{41} & \textcolor{black}{0}\\
        ChpperCmmnd & - & \textcolor{red}{60} & - & \textcolor{red}{60}\\
        DemonAttack & \textcolor{Green}{-1} & \textcolor{Green}{-4} & \textcolor{red}{1} & \textcolor{Green}{-4}\\
        Hero & \textcolor{red}{1} & \textcolor{red}{5} & \textcolor{red}{10} & \textcolor{red}{5}\\
        Phoenix & \textcolor{Green}{-10} & \textcolor{Green}{-10} & \textcolor{Green}{-26} & \textcolor{Green}{-10}\\
        Pong & \textcolor{red}{19} & \textcolor{red}{20} & \textcolor{red}{54} & \textcolor{red}{20}\\
        Qbert & \textcolor{red}{Failed} & \textcolor{black}{0} & \textcolor{red}{Failed} & \textcolor{black}{0}\\
        Riverraid & \textcolor{red}{Failed} & \textcolor{red}{2} & \textcolor{red}{Failed} & \textcolor{red}{2}\\
        RoadRunner & - & \textcolor{red}{Failed} & - & \textcolor{red}{Failed}\\
        SpaceInvaders & \textcolor{red}{21} & \textcolor{red}{7} & \textcolor{red}{2} & \textcolor{red}{7}\\
        StarGunner & \textcolor{black}{0} & \textcolor{black}{0} & \textcolor{Green}{-1} & \textcolor{black}{0}\\
		\bottomrule
    \end{tabularx}
    \normalsize
\end{table*}

\subsection{Producing pruned policy graphs}

We evaluate the pruned policies as we progressively 
increase the number of states in which we take the 
original policy over the default action. We then 
plot the performance of the evaluated policies, with 
respect to either the percentage of states restored, or
the percentage of steps in which the original policy 
is used.

\paragraph{Encountering previously unvisited states in the pruned policy experiments}
In our pruned policy experiments (Figure~\ref{fig:graphs} and Table~\ref{tab:pruning}),
the agent may encounter states that were not seen during the ranking procedure 
(and therefore not ranked). In this
case, we have the agent use the default action, rather than the original policy.
This means that in the graphs, at the point where Original States (\%) is 100\%, the agent
will still take the default action if it encounters a state that was 
never encountered during the ranking procedure. As a result, it is 
possible that the pruned policy never recovers the performance of the original 
policy. The Atari game Boxing is an especially extreme example of this, where
only 30\% of the original performance is recovered. This is because 75\% of the 
states being visited had not been seen during the ranking procedure, meaning that 
the agent was mostly taking default actions. To avoid this, we could (1) run the 
ranking procedure for longer, (2) use a state abstraction method to reduce the size 
of the state space we are dealing with, or (3) set the pruned policy such that it
uses the original policy in any unvisited state, rather than the default action.

\paragraph{Pruned policies do not improve monotonically}
Because of the complex 
interactions between the states,
the policies do not monotonically improve as 
states are restored. For example,
restoring a state could send the agent down a new trajectory
in which other states have not yet been restored, and
so the agent may fail. As a result, we show in each 
point the performance achievable for that percentage 
(of either states restored or policy's actions taken)
\textit{or less}. Suppose a policy $q$ recovers 
60\% of the original policy's reward after having 
restored 35\% of states. If there exists another policy 
which recovers 70\% of reward with only 30\% 
of states, then it is strictly better, and 
we should use this policy over $q$. As a result, 
we report having recovered 70\% of the reward 
at both 30\% and 35\% of restored states.

\paragraph{SBFL portfolio ranking}
We present the SBFL portfolio ranking as 
the combination of the different SBFL measures
presented. As we are free to use any of the measures,
we use the best measure in each point.
The SBFL curves show the best performance for any 
of the measures in each point.

\paragraph{Normalising scores}
For Minigrid and CartPole, the minimum score is 
0, and so we simply show what percentage the pruned policy 
achieves of the original policy's reward. In Atari games,
it is common to normalise reward with respect to the 
reward of a random policy. We use the figures 
in Table~\ref{tab:ataripol} to compute 
$100\times$(pruned~score~-~random score)~/~(original~score~-~random~score).

\paragraph{Producing data for ``Policy's Action Taken (\% of Steps)''}
We cannot directly control the \% of steps in which the agent will use 
the original policy, because we do not know what the entire trajectory will
be a priori (and so how many steps it will be). Instead, we use the same 
data as for the experiments where we use the original policy in the
top-x \% of states. For each data-point, we calculate what fraction of the 
actions taken were from the original policy, rather than the default action. 
We are then able to use this transformed data.

\subsection{Low-performance policies on CartPole}
Figure~\ref{fig:cpolegoodbad} expands on results 
for \textbf{RQ4}. We use the rankings based on 
scoring states of the good policy using FreqVis and 
Tarantula. For each of these two rankings, we are able 
to look at the agreement between the two policies 
for the top X\% of states, as X increases. The overall 
agreement in all states considered is about 64\%. 
The results show that the policies disagree greatly 
in the 10\% most frequently visited states, but beyond
this frequency of visit is a poor predictor of agreement.
On the other hand, we find that the policies agree 
in most top-ranked tarantula states.

\begin{figure}[t]
	\centering 
	\includegraphics[width=0.7\linewidth]{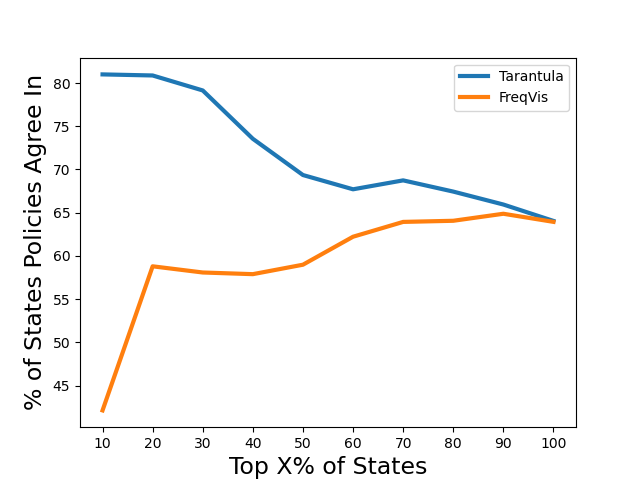}
	\caption{In CartPole, of the top X\% of states
	ranked by either Tarantula or FreqVis, 
	the percentage of agreement between a low and 
	high-quality policy. Data taken from 1,000 runs
	through the environment.}
	\label{fig:cpolegoodbad}
\end{figure}
}{}

\end{document}

